\pdfoutput=1

\documentclass[mlmain]{jmlr}

\usepackage{longtable}
\usepackage{booktabs}
\usepackage[load-configurations=version-1]{siunitx} % 

\newcommand{\subsubsectioninline}[1]{\vspace{0.5em}\noindent\textbf{#1}}

\newcommand{\todoc}[2][]{\todo[color=Apricot!10,#1]{#2}}
\newcommand{\todoar}[2][]{\todo[color=Blue!30,#1]{#2}}
\renewcommand{\todoc}[2][]{}
\renewcommand{\todoar}[2][]{}

\def\E{{\mathbb E}}

\def\E{{\mathbb E}}

\def\beqa{\begin{eqnarray}}
\def\eeqa{\end{eqnarray}}
\def\beqann{\begin{eqnarray*}}
\def\eeqann{\end{eqnarray*}}

\renewcommand{\epsilon}{\varepsilon}

\title{Random Field Augmentations for Self-Supervised Representation Learning}

\author{\Name{Philip Andrew Mansfield\nametag{\thanks{Equal contribution}}} \Email{memes@google.com}\\
\addr Google Research
\AND
\Name{Arash Afkanpour\nametag{\footnotemark[\value{mpfootnote}]}} \Email{arash.afkanpour@vectorinstitute.ai}\\
\addr Vector Institute
\AND
\Name{Warren Richard Morningstar} \Email{wmorning@google.com}\\
\addr Google Research
\AND
\Name{Karan Singhal} \Email{karansinghal@google.com}\\
\addr Google Research
}

\begin{document}

\maketitle

\begin{abstract}
Self-supervised representation learning is heavily dependent on data augmentations to specify the invariances encoded in representations. Previous work has shown that applying diverse data augmentations is crucial to downstream performance, but augmentation techniques remain under-explored. In this work, we propose a new family of local transformations based on Gaussian random fields to generate image augmentations for self-supervised representation learning. These transformations generalize the well-established affine and color transformations (translation, rotation, color jitter, etc.) and greatly increase the space of augmentations by allowing transformation parameter values to vary from pixel to pixel. The parameters are treated as continuous functions of spatial coordinates, and modeled as independent Gaussian random fields. Empirical results show the effectiveness of the new transformations for self-supervised representation learning. Specifically, we achieve a 1.7\% top-1 accuracy improvement over baseline on ImageNet downstream classification, and a 3.6\% improvement on out-of-distribution iNaturalist downstream classification. However, due to the flexibility of the new transformations, learned representations are sensitive to hyperparameters. While mild transformations improve representations, we observe that strong transformations can degrade the structure of an image, indicating that balancing the diversity and strength of augmentations is important for improving generalization of learned representations. 
\end{abstract}

\begin{keywords}
Self-supervised learning, Representation learning, Gaussian random fields, Local symmetry
\end{keywords}

\section{Introduction}
\label{sec:introduction}

Data augmentations play a crucial role in joint embedding self-supervised representation learning methods. They specify the transformations under which the representations must remain invariant. In the absence of any prior knowledge, most self-supervised learning methods assume that each data point is semantically different from other examples in the data set. Data augmentations, on the other hand, relate each example to its transformed versions via a soft positive label. 
While some previous work studied the impact of augmentations on representations \citep{chen2020simple, caron2020unsupervised}, for the most part this remains an under-explored area in self-supervised learning. Since these transformations specify what representations learn, a natural question is whether additional diverse transformations improve generalizability and robustness of representations.

In this work we introduce and study a family of visual local transformations based on \emph{Gaussian random fields}. In particular we define local spatial and color transformations to modify the position and color of pixels using Gaussian random fields. The new transformations are a generalization of the standard affine (rotation, translation, etc.) and color transformations used in many methods \citep{chen2020simple,grill2020bootstrap,chen2021exploring} and operate at the pixel level. Our empirical results in both in-distribution and out-of-distribution tasks demonstrate the effectiveness of these transformations for representation learning.

\section{Related work}
\label{sec:related_work}

\subsection{Joint Embedding Methods}
Joint embedding self-supervised learning methods use a variety of objective functions to create invariance of representations across multiple views of the same images. These views are usually generated by applying several transformations that do not change the semantics of an image. Based on the objective function, these methods can be divided into several categories. For example, contrastive methods such as CPC \citep{oord2018representation}, SimCLR \citep{chen2020simple} and MoCo \citep{he2020momentum} use InfoNCE contrastive loss to pull representations of different augmentations of an image together, while pushing representations of different images apart. Clustering methods, e.g., DeepCluster \citep{caron2018deep} and SwAV \citep{caron2020unsupervised} use a combination of clustering and constrastive loss to learn a similar representation for different views of an image. Canonical correlation analysis methods, such as Barlow Twins \citep{zbontar2021barlow} and VICReg \citep{bordes2023towards} rely on correlation analysis of features in the representation space. Their objective is defined to maximize correlation of the same feature across multiple views, while decorrelating different features. Self-distillation methods such as BYOL \citep{grill2020bootstrap} and SimSiam \citep{chen2021exploring} use a dual encoder architecture where one encoder is a slightly different version of the other (e.g., an exponential moving average encoder in BYOL). The model is trained by maximizing similarity between representations of the encoders fed with two views of the same image.

In contrast to joint embedding methods, representation learning based on masked image modeling does not rely on data augmentations. Similar to the masked token prediction task in BERT pretraining \citep{devlin2018bert} the general principle is to mask parts of an image and minimize a loss to reconstruct them given the remaining parts. Most notably \citet{he2022masked} takes advantage of vision transformers \citep{dosovitskiy2020image} to learn representations with this approach.

\subsection{Image Augmentations}
SimCLR \citep{chen2020simple} studied the effectiveness of several augmentations including random crop, cutout, color jitter, Sobel filter, Gaussian blur, Gaussian noise, and global rotation. They examined individual and pairs of augmentations for representation learning. They observed that random crop and color jitter are the most effective augmentations when these representations are used for ImageNet classification. Most subsequent work in self-supervised representation learning, e.g. \citet{chen2020big, zbontar2021barlow, bardes2021vicreg}, use the same set of augmentations. One exception is multi-crop proposed by \citet{caron2020unsupervised} where multiple small crops are taken as additional views of a source image. In this case the model is trained to produce the same representation for small crops and views generated by the composition of other augmentations.

\citet{bordes2023towards} studied the impact of different combinations of augmentations. They showed the combination of random crop and a grayscale transformation is quite competitive, measured by classification accuracy on ImageNet, to the full augmentation set.

One of the shortcomings of the current augmentations is that they are selected to achieve the best performance on ImageNet classification. It is possible that representations learned via these augmentations do not perform well on other downstream tasks. \citet{ericsson2021self} investigated how the learned invariances affect the performance across a diverse set of downstream tasks. They showed that in some tasks a subset of data augmentations outperforms the default combination of SimCLR augmentations.

Image augmentations remain under-explored given their importance to representation learning, especially for out-of-distribution downstream tasks, motivating our work.

\section{Random Field Transformations}
\label{sec:grf_transformation}

\subsection{Gaussian Random Fields}

A local transformation is characterized by one or more parameter fields where each (pixel) position has its own transformation parameter(s). A random parameter field ensures diversity of transformations. At the same time, complete independence of parameters results in distortions that make the final image unrecognizable. Therefore parameters must be relatively slowly varying continuous functions of spatial coordinates, and nearby values of the random field must be suitably correlated with each other. Gaussian random fields offer a convenient mathematical tool for this purpose. Here we provide a brief description of Gaussian random fields. There are numerous resources on this topic, cf. \citet{adler2007random}.

A random field is a stochastic process with a structured parameter space. Let $\mathcal{X}$ denote a parameter space, such as the Euclidean space. Given $\mathcal{X}$, a random field $\phi$ is a collection of random variables
\beqann
\{ \phi(x): x \in \mathcal{X} \}.
\eeqann
In a Gaussian random field any finite number of variables constitute a multivariate Gaussian distribution. Therefore, a Gaussian random field is fully characterized by its mean ($\mu$) and covariance ($\Sigma$) functions:
\beqann
    \mu(x) &=& \E [\phi(x)], \\
    \Sigma(x, y) &=& \E [(\phi(x) - \mu(x))(\phi(y) - \mu(y))].
\eeqann
If the mean of a random field is constant across $\mathcal{X}$ and the covariance is a function of the difference $(x - y)$ only, then the random field is \emph{homogeneous}. Additionally, if the covariance is a function of the Euclidean distance $|x - y|$ then $\phi$ is also \emph{isotropic}. With some abuse of notation an isotropic random field is usually written as $\Sigma(x, y) = \Sigma(|x - y|)$. A homogenous and isotropic Gaussian random field is particularly interesting because it is fully characterized by its covariance (equivalently correlation) function, and this function only depends on the distance between points in the parameter space.

\begin{figure}[t]
     \centering
     \includegraphics[width=0.99\textwidth]{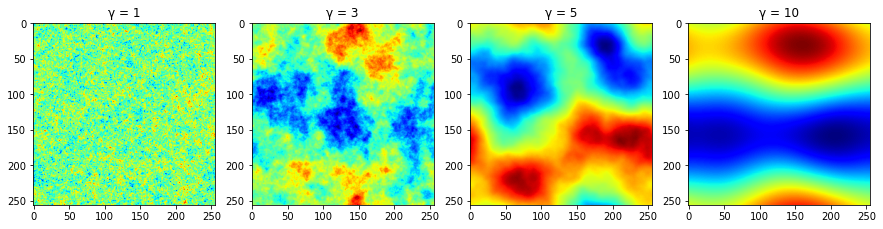}
     \caption{Gaussian random fields with different values of the power law exponent.}
     \label{fig:random_fields}
\end{figure}

Generating a random field in the spatial domain is computationally expensive. However, it can be easily calculated in the frequency domain. The power spectrum, which is the Fourier transform of the correlation function, characterizes a Gaussian random field in the frequency domain. In our experiments we specified the power spectrum as \emph{power law}: $P(k) \propto k^{- \gamma}$, where $\gamma$ controls the correlation of points in the spatial domain: larger values result in higher correlation among distant points. Figure~\ref{fig:random_fields} shows examples of random fields with different $\gamma$ values.

\subsection{Image Transformations with Gaussian Random Fields}
Spatial affine transformations such as rotation, translation, scaling, etc. are usually parameterized by a few parameters that specify the magnitude of transformation globally. Consider the translation transformation. It requires two parameters, $t_X$ and $t_Y$ which determine the amount of translation across X and Y axes respectively. One way to generalize this transformation is to use pixel-specific translation values, i.e. $t_X(x, y)$ and $t_Y(x, y)$, where $t_X$ and $t_Y$ are Gaussian random fields. To ensure images remain recognizable, transformations are set up such that local changes are small. This is primarily controlled by $\gamma$, the exponent of the power law used as the spectrum function. We loosely use the term \emph{kernel width} to refer to this parameter. A large value for kernel width indicates a strong correlation between pixels even if they are far apart, resulting in a smoother random field. In addition, we limit the magnitude of the random field by a parameter $\alpha$ such that $-\alpha \leq \theta(x, y) \leq \alpha$, where $\theta$ denotes the random field. Eq.~\ref{eq:local_affine_transform} shows the general form of a local affine transformation applied to a 2-dimensional source point $(x^s, y^s)$.
\beqa
\begin{bmatrix}
x^s \\
y^s
\end{bmatrix} = 
\begin{bmatrix}
\theta_{11}(x, y) & \theta_{12}(x, y) & \theta_{13}(x, y) \\
\theta_{21}(x, y) & \theta_{22}(x, y) & \theta_{23}(x, y)
\end{bmatrix} 
\begin{bmatrix}
x^t \\
y^t \\
1
\end{bmatrix}
\label{eq:local_affine_transform}
\eeqa
As is common in Computer Graphics, we multiply the transformation matrix by the target coordinates, $(x^t, y^t)$, to fully cover the target space. Multiplication by the source coordinates on the other hand could result in undefined values for some target coordinates. See more details in \citet{foley1994introduction}.

In our experiments we focused on four common affine transformations: rotation, scaling, shearing, and translation. For example, the local scale transformation matrix is given by,
\beqa
\theta_{scale}(x, y; \gamma_x, \gamma_y, \alpha_x, \alpha_y) =
\begin{bmatrix}
1 + g_x(x, y; \gamma_x, \alpha_x) & 0 & 0 \\
0 & 1 + g_y(x, y; \gamma_y, \alpha_y) & 0
\end{bmatrix} 
\label{eq:local_scale}
\eeqa
where $g_x$ and $g_y$ are independent Gaussian random fields, parameterized by smoothness parameters $\gamma$ and scale factors $\alpha$ such that $-\alpha \leq g(x, y; \gamma, \alpha) \leq \alpha$. In Eq.~\ref{eq:local_scale} we use different random fields for the $X$ and $Y$ axes.
The matrices of other affine transformations are available in Appendix~\ref{app:local_transformation_matrices}.

\begin{figure}[t]
    \centering
    \begin{subfigure}
        \centering
        \includegraphics[width=0.95\textwidth]{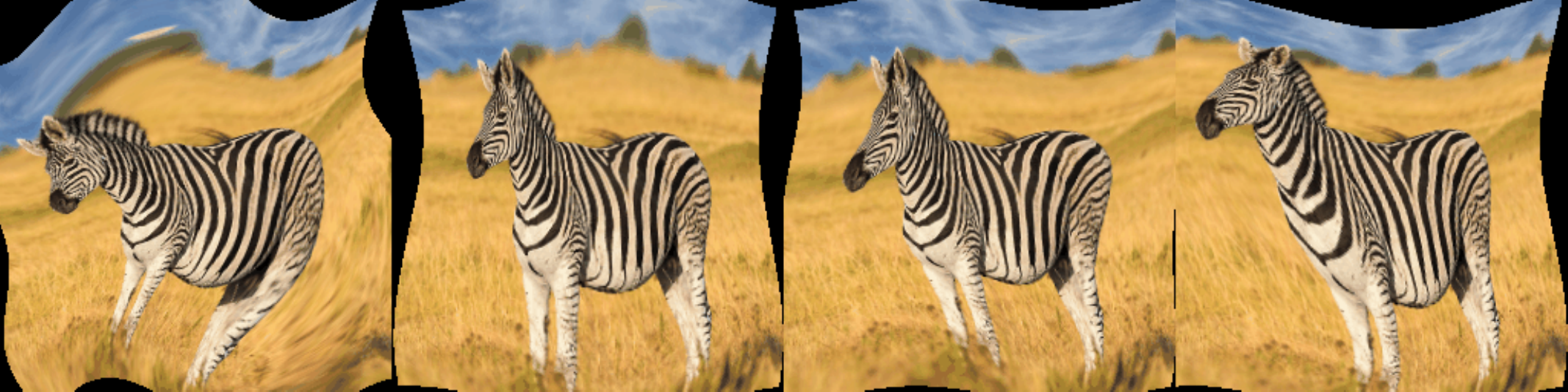}
    \end{subfigure}
    \begin{subfigure}
        \centering
        \includegraphics[width=0.95\textwidth]{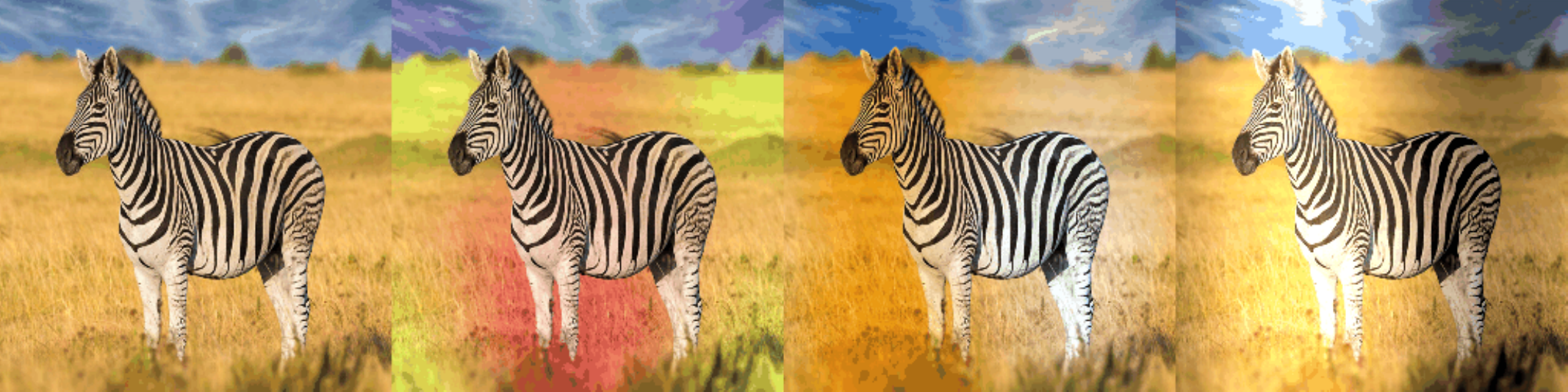}
    \end{subfigure}
    \caption{Local transformations with Gaussian random fields. Top left to top right: rotation, scale, shear, translation. Bottom left to bottom right: original, hue, saturation, brightness.}
    \label{fig:local_transformation_examples}
\end{figure}

We apply local color transformations to hue, saturation, and value channels separately. For each channel, a Gaussian random field is added to the channel values to obtain the new values. Figure~\ref{fig:local_transformation_examples} shows examples of local affine and color transformations.

\section{Empirical Results}
\label{sec:results}

In all experiments we use SimCLR \citep{chen2020simple} as the self-supervised representation learning method. Pretraining of the encoder is performed on the ImageNet training split with 1.2 million images. Following the linear probing protocol of previous papers, we evaluated each setting by training a linear classifier on the output representations of the frozen encoder network. Our downstream tasks are image classification on two datasets: ImageNet (in-distribution) and iNaturalist 2018 (out-of-distribution). In each case a linear classifier was trained on the training split of the dataset and then evaluated on the validation split.

In all experiments we apply local transformations in addition to the standard SimCLR augmentations \citep{chen2020simple}. Random field augmentations are applied before the SimCLR augmentations, with the exception of crop and resize, which we apply as the first augmentation to resize images to 224$\times$224 in order to reduce the computational cost of local transformations.

\subsection{Atomic Local Transformations}
In this experiment we evaluate five local transformations: color jitter, rotation, scaling, shearing, and translation. We choose each parameter range so that the resulting transformation does not make the images unrecognizable. For each image, $\gamma$ (the random field smoothness parameter) is sampled uniformly from $[7, 10]$. The random field scale factor ($\alpha$) is uniformly sampled from $[0, 1/3]$. A local transformation is applied to each of the two views of SimCLR with probability $0.8$. Table~\ref{tab:single_transformations} shows Top-1 and Top-5 classification accuracy on the ImageNet and iNaturalist 2018 downstream tasks.

In the ImageNet task (in-distribution) local scale, shear, and translate outperform the baseline. Local color jitter and rotation, on the other hand degrade accuracy. In the iNaturalist task (out-of-distribution) all local transformations outperform the baseline. In both cases local color jitter slightly underperforms local affine transformations, which could indicate that the classification task is more sensitive to local color changes than local spatial changes. Another observation is that local rotation performance is generally slightly worse than other local affine transformations. This could be due to larger structural changes made to an image by local rotation compared to other local affine transformations (see Figure~\ref{fig:local_transformation_examples} for an example).

\begin{table}[t]
\centering
\begin{tabular}{c|c|c}
\toprule
 & \textbf{ImageNet} & \textbf{iNaturalist} \\
 & \textbf{Top-1 / Top-5} & \textbf{Top-1 / Top-5} \\
\midrule
Baseline (SimCLR augmentations) & 0.7056 / 0.9022 & 0.3873 / 0.5983 \\
Local color jitter & 0.7045 / 0.9013 & 0.3964 / 0.6071 \\
Local rotate & 0.7007 / 0.8945 & 0.4159 / 0.6171 \\
Local scale & 0.7102 / 0.8964 & 0.4174 / 0.6245 \\
Local shear & 0.7219 / 0.9031 & 0.4102 / 0.6228 \\
Local translate & \textbf{0.7223 / 0.9015} & \textbf{0.4231 / 0.6267} \\
\bottomrule
\end{tabular}
\caption{Effect of atomic random field augmentations (in addition to SimCLR augmentations) on learned representations measured by downstream classification accuracy. Bold numbers indicate the highest Top-1 accuracy.}
\label{tab:single_transformations}
\end{table}

\subsection{Effect of Random Field Parameters}
\label{sec:parameter_sweep}
We performed a grid search on the two parameters of the random fields, $\gamma$ and $\alpha$. For each parameter we specified different intervals for uniform sampling. The $\gamma$ intervals include $[3, 7]$, $[7, 10]$, and $[3, 10]$. Usually $\gamma < 3$ yields strong local distortions that destroy the global structure, making an image unrecognizable. On the other end $\gamma > 10$ yields almost no difference to augmentations with $\gamma = 10$. The $\alpha$ intervals include $[0, 1/3]$, $[0, 2/3]$, and $[0, 1]$. In this experiment we focus on local translate and apply it to both views of SimCLR, each with probability $0.8$. Similar to the previous experiment, we follow the standard linear probing protocol by training a linear classifier on the output of a frozen encoder. Table~\ref{tab:parameter_sweep} shows top-1 and top-5 accuracy numbers on each validation set.

Among these combinations $\gamma \in [7, 10], \alpha \in [0, 1/3]$ leads to the best classification accuracy on both downstream tasks. Strong distortions, achieved by smaller $\gamma$ or larger $\alpha$ could lead to transformations that change the spatial structure of images too drastically, leading to worse performance of representations in downstream tasks.

\begin{table}[tb]
\centering
\begin{tabular}{c|c|c|c}
\toprule
 \textbf{ImageNet} & $\alpha \in [0, 1/3]$ & $\alpha \in [0, 2/3]$ & $\alpha \in [0, 1]$ \\
\midrule
$\gamma \in [3, 7]$  & 0.6981 / 0.8896 & 0.6879 / 0.8788 & 0.6595 / 0.865 \\
$\gamma \in [7, 10]$ & \textbf{0.7223 / 0.9015} & 0.6939 / 0.8873 & 0.6917 / 0.8879 \\
$\gamma \in [3, 10]$ & 0.7045 / 0.8939 & 0.6937 / 0.8808 & 0.6723 / 0.8697 \\
\bottomrule
\end{tabular}
\begin{tabular}{c|c|c|c}
\toprule
 \textbf{iNaturalist} & $\alpha \in [0, 1/3]$ & $\alpha \in [0, 2/3]$ & $\alpha \in [0, 1]$ \\
\midrule
$\gamma \in [3, 7]$  & 0.4046 / 0.6055 & 0.4043 / 0.6021 & 0.3964 / 0.5895 \\
$\gamma \in [7, 10]$ & \textbf{0.4231 / 0.6267} & 0.4080 / 0.6169 & 0.4180 / 0.6245 \\
$\gamma \in [3, 10]$ & 0.4183 / 0.6255 & 0.4136 / 0.6092 & 0.4043 / 0.6003 \\
\bottomrule
\end{tabular}
\caption{Top-1 / Top-5 classification accuracy of representations trained with local translation over different ranges of $\gamma$ and $\alpha$. Top: ImageNet, bottom: iNaturalist 2018.}
\label{tab:parameter_sweep}
\end{table}

With the best combination of parameters, i.e. $\gamma \in [7, 10], \alpha \in [0, 1/3]$, we performed a sweep over the probability parameter that determines how often a random field transformation is applied to the image. Table~\ref{tab:probability_sweep} shows the results. Broadly, as the probability value increases, downstream classification accuracy increases too. This trend continues until reaching maximum accuracy at $p=0.8$. Pushing the probability value further to 1.0, however, leads to a decline in accuracy, similar to other work that has observed benefit to applying augmentations stochastically.

\begin{table}[t]
\centering
\begin{tabular}{c|l|l}
\toprule
 \textbf{Probability} & \textbf{ImageNet} & \textbf{iNaturalist} \\
 & \textbf{Top-1 / Top-5} & \textbf{Top-1 / Top-5} \\
\midrule
0.0 & 0.7056 / 0.9022 & 0.3873 / 0.5983 \\
0.2 & 0.7134 / 0.8971 & 0.3906 / 0.6030 \\
0.4 & 0.7207 / 0.9034 & 0.3954 / 0.6060 \\
0.6 & 0.7140 / 0.9036 & 0.3903 / 0.6060 \\
0.8 & 0.7223 / 0.9015 & 0.4231 / 0.6267 \\
1.0 & 0.6929 / 0.8893 & 0.3937 / 0.6031 \\
\bottomrule
\end{tabular}
\caption{Top-1 / Top-5 classification accuracy of downstream classification for different values of the probability parameter.}
\label{tab:probability_sweep}
\end{table}

\subsection{Composite Transformations}

We study composite affine transformations in this section. For simplicity we only consider the composition of two atomic local transformations. For each atomic transformation $\gamma$ and $\alpha$ are sampled uniformly from $[7, 10]$, and $[0, 1/3]$ respectively. In order to ensure that the combination of transformations remain within the same bounds as individual transformations the scale factor of each transformation is multiplied by $1 / \sqrt{2}$ before application.\footnote{To combine $N$ transformations, this coefficient should be $1 / \sqrt{N}$.} A composite transformation is then formed by multiplying the matrices of individual transformations in random order. Similar to the previous experiments each composite transformation is applied with probability $0.8$. Table~\ref{tab:composite_transformations} shows the results. While in the ImageNet task (in-distribution) combining transformations generally improves performance, in the iNaturalist task (out-of-distribution) performance degrades by combining local transformations. Since combining local transformations can generally be interpreted as stronger distortions in the local structure of an image, these results indicate that too strong distortions could have a negative effect on representations. This observation is also supported by the results in Section~\ref{sec:parameter_sweep}.

\begin{table}[t]
\small
\centering
\begin{tabular}{c|c|c|c|c}
\toprule
\textbf{ImageNet} & Rotate & Scale & Shear & Translate  \\ \hline
Rotate & 0.7007 & 0.7092 & 0.7155 & 0.716 \\ \hline
Scale & - & 0.7102 & 0.7235 & 0.7088 \\ \hline
Shear & - & - & 0.7219 & 0.7119 \\ \hline
Translate & - & - & - & 0.7223 \\
\bottomrule
\end{tabular}
\hspace{2mm}
\begin{tabular}{c|c|c|c|c}
\toprule
\textbf{iNaturalist} & Rotate & Scale & Shear & Translate \\ \hline
Rotate & 0.4159 & 0.4026 & 0.4006 & 0.4044 \\ \hline
Scale & - & 0.4174 & 0.3848 & 0.3953 \\ \hline
Shear & - & - & 0.4102 & 0.3966 \\ \hline
Translate & - & - & - & 0.4231 \\
\bottomrule
\end{tabular}
\caption{Top-1 downstream classification accuracy of composite transformations on ImageNet (top) and iNaturalist (bottom) data sets. Diagonal elements show the accuracy of atomic transformations.}
\label{tab:composite_transformations}
\end{table}

\section{Conclusion}
\label{sec:conclusion}

Image augmentations play a crucial role in joint embedding self-supervised learning methods. Yet different augmentation methods have been studied minimally in this context. This motivates work exploring whether additional diverse augmentations could result in more robust and generalizable representations. In this paper we introduced random field augmentations as a generalization of some of the previous forms of augmentations, in particular crop-and-resize (equivalently scale and translate) and color jitter, which according to \citet{chen2020simple} are the two most effective augmentations for SimCLR. Our new transformations vastly increase the space of augmentations by enabling coordinate-based transformations where transformation parameters are selected according to Gaussian random fields.

We performed multiple empirical studies for in-distribution and out-of-distribution cases. These studies include a comparison of different types of transformations, measuring the effect of transformation parameters on the quality of representations and a comparison of composite transformations. The results showed effectiveness of the new transformations when applied in addition to the standard transformations of SimCLR. Due to the flexibility of the new transformations, careful hyperparameter tuning must be performed on the random field parameters. While mild transformations generally improve representations, we showed that strong transformations, which could significantly change the structure of an image, led to performance degradation. 

Future work can apply random field augmentations to different self-supervised representation learning methods with different model architectures and downstream tasks, studying the effect of these flexible transformations on generalization in different contexts. 
\bibliography{references}

\appendix

\section{Local Transformation Matrices}
\label{app:local_transformation_matrices}

In all transformations the center of the coordinate system is the center of the image.
Let $g(x, y; \gamma, \alpha)$, $g_x(x, y; \gamma_x, \alpha_x)$ and $g_y(x, y; \gamma_y, \alpha_y)$ be Gaussian random fields. The atomic local affine transformations are defined as follows:

\subsubsectioninline{Local Rotate}
\beqann
\theta_{rotate}(g) =
\begin{bmatrix}
\cos{\pi g} & -\sin{\pi g} & 0 \\
\sin{\pi g} & \cos{\pi g} & 0
\end{bmatrix} 
\eeqann

\subsubsectioninline{Local Scale}
\beqann
\theta_{scale}(g_x, g_y) =
\begin{bmatrix}
1 + g_x & 0 & 0 \\
0 & 1 + g_y & 0
\end{bmatrix} 
\eeqann

\subsubsectioninline{Local Shear}
\beqann
\theta_{shear}(g_x, g_y) =
\begin{bmatrix}
1 & g_x & 0 \\
g_y & 1 & 0
\end{bmatrix} 
\eeqann

\subsubsectioninline{Local Translate}
\beqann
\theta_{translate}(g_x, g_y) =
\begin{bmatrix}
1 & 0 & g_x \\
0 & 1 & g_y
\end{bmatrix} 
\eeqann

\end{document}